\documentclass{article} % For LaTeX2e
\usepackage{iclr2025_conference,times}
\iclrfinalcopy

\usepackage{amsmath,amsfonts,bm}

\def\eqref#1{equation~\ref{#1}}
\def\1{\bm{1}}

\DeclareMathAlphabet{\mathsfit}{\encodingdefault}{\sfdefault}{m}{sl}
\SetMathAlphabet{\mathsfit}{bold}{\encodingdefault}{\sfdefault}{bx}{n}

\usepackage{hyperref}
\usepackage{url}
\usepackage{graphicx}

\usepackage{colortbl}
\usepackage{tabulary}
\usepackage{makecell}
\usepackage{threeparttable}
\usepackage{wrapfig}

\usepackage{movie15}
\usepackage{animate}

\usepackage{amsmath}
\usepackage{multirow}
\usepackage{booktabs} 
\usepackage{subcaption} 
\usepackage{amsfonts}       % blackboard math symbols
\usepackage{pifont}% http://ctan.org/pkg/pifont
\newcommand{\cmark}{\ding{51}}%
\newcommand{\xmark}{\ding{55}}%

\newcommand{\methodname}{\textsc{Unbounded}}

\title{\textbf{Unbounded}: A Generative Infinite Game of Character Life Simulation}

\author{Jialu Li$^{1,2}$\thanks{Work done during an internship at Google} \;\;\;\;
Yuanzhen Li$^{1}$\;\;\;\;
Neal Wadhwa$^{1}$\;\;\;\;
Yael Pritch$^{1}$\;\;\;\;
David E. Jacobs$^{1}$\;\;\;\; \\
\textbf{Michael Rubinstein}$^{1}$\;\;\;\; 
\textbf{Mohit Bansal}$^{2}$\;\;\;\;
\textbf{Nataniel Ruiz}$^{1}$  \\
$^{1}$Google\;\;\;\;   $^{2}$The University of North Carolina at Chapel Hill \\
\\
\;\;\;\;\;\;\;\;\;\;\;\;\;\;\;\;\;\;\;\;\;\;\;\;\;\;\;\;\;\;\;\;\;\;\;\;\;\;\;\;\textbf{\textcolor{blue}
{\href{https://generative-infinite-game.github.io/}{generative-infinite-game.github.io}}}
}

\begin{document}

\maketitle

\begin{figure}[ht]
    \centering
    \vspace{-0.1in}
    \includegraphics[width=0.9\textwidth]{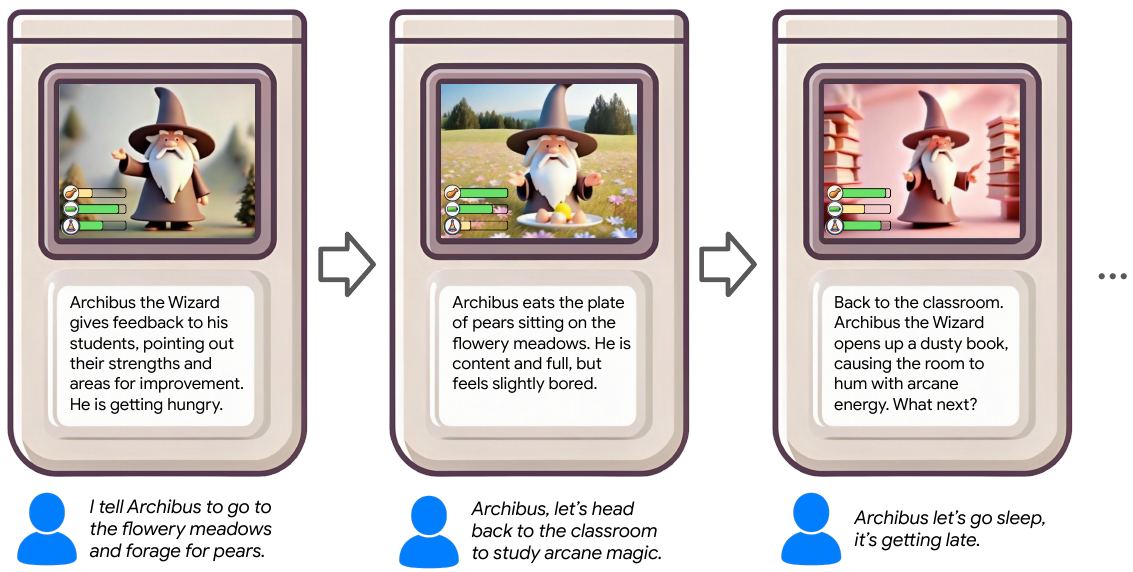}
    \vspace{-5pt}
    \caption{
    An example of \methodname{}. We follow the life of \textit{Archibus}, the user's custom wizard character. The user can interact with the generative game using natural language, and Archibus' hunger, energy and fun meters update accordingly. A spontaneous and unconstrained story unfolds while the user plays, and the character can explore new environments with a myriad of possible actions and unexpected interactions. The game runs in interactive speeds, refreshing every second.
    }
    \label{fig:teaser}
    \vspace{-0.05in}
\end{figure}

\begin{abstract}
We introduce the concept of a \textbf{generative infinite game}, a video game that transcends the traditional boundaries of finite, hard-coded systems by using generative models. Inspired by James P. Carse's distinction between finite and infinite games~\citep{carse1986finite}, we leverage recent advances in generative AI to create \textbf{\methodname{}}: a game of character life simulation that is fully encapsulated in generative models. Specifically, \methodname{} draws inspiration from sandbox life simulations and allows you to interact with your autonomous virtual character in a virtual world by feeding, playing with and guiding it - with open-ended mechanics generated by an LLM, some of which can be emergent. In order to develop \methodname{}, we propose technical innovations in both the LLM and visual generation domains. Specifically, we present: (1) a specialized, distilled large language model (LLM) that dynamically generates game mechanics, narratives, and character interactions in real-time, and (2) a new dynamic regional image prompt Adapter (IP-Adapter) for vision models that ensures consistent yet flexible visual generation of a character across multiple environments. We evaluate our system through both qualitative and quantitative analysis, showing significant improvements in character life simulation, user instruction following, narrative coherence, and visual consistency for both characters and the environments compared to traditional related approaches.
\end{abstract}

\section{Introduction}

In his work \textit{``Finite and Infinite Games: A Vision of Life as Play and Possibility''}~\citep{carse1986finite}, James P. Carse introduces a distinction between two types of games. Carse defines \textbf{finite games} as those \textit{``played for the purpose of winning,''} with boundaries, fixed rules, and a definitive endpoint. In contrast, \textbf{infinite games} are \textit{``played for the purpose of continuing the play,''} with no fixed boundaries and evolving rules. Traditional video games are, inherently, finite games due to the limitations of programming and computer graphics. For example, game mechanics have to be fully pre-defined in the programming language and graphics assets usually have to be pre-designed (modulo traditional procedural generation which still grapples with structural limitations). This allows for only a finite, and sometimes predefined, set of actions and paths that can be taken. They also feature predefined rules, boundaries, and win conditions, which run counter to the infinite game definition.

Recent advances in generative models have been impressive. We hypothesize that these developments have finally opened up the possibility of creating the first generative infinite video game - an infinite game that is also fully subsumed in generative models with no logic or graphics being controlled by other, more traditional, processes. Two advancements have made this achievable: (1) large language models (LLMs) that can encode persistent video game mechanics (e.g. interactions with the game environment or characters, character state tracking, object permanence), generate interactive stories, and produce spontaneous, and sometimes emergent, behaviors; and (2) visual generative models capable of producing high-quality images that follow prompts. In this work, we present \textbf{\methodname{}}, what we believe to be the first interactive generative infinite game, where all game behaviors and graphics are generated by AI models, transcending the constraints of hard-coded systems. \methodname{} draws inspiration from sandbox life simulations and digital pet games such as Little Computer People, The Sims and Tamagotchi. It incorporates elements from tabletop roleplaying games like Dungeons \& Dragons, which offer unconstrained storytelling experiences that have been unattainable in video games.

Some useful prior work has proposed ideas that are related to our concept of generative infinite games. \cite{gaudl2018exploring} proposes fluidic games that merge gameplay with game design, automated game design~\citep{cook2016angelina} explores automated systems that can help in making new games, and work on AI-based games~\citep{treanor2015aibased} proposes design techniques for games that put AI at the forefront of user experience. These concepts are different from our definition of generative infinite games since they emphasize structured exploration within predefined game parameters and tools, where the design space is extensive yet ultimately bounded. In contrast, infinite games, as we envision them, aim to continuously evolve beyond predefined structures, with generative models dynamically creating both content and mechanics in real-time, enabling an open-ended gameplay experience with no fixed endpoint or limitations on possible interactions. There is also important work on using machine learning models as components that generate parts of a game. \cite{agarwal2024infinite} generates novel interactions, \cite{sun2023language} generates new mechanics and items as well as some graphical assets, and there is also work on orchestration of different models for game design~\citep{liapis2024orchestrating}. These works do not generate all components of the game. In contrast, in our work, all of the game mechanics, characters, environments, narrative, and graphics are fully produced by generative models. This is more similar in spirit to recent work Genie~\citep{bruce2024genie} and GameNGen~\citep{valevski2024diffusion}, although in contrast to these works that mainly generate platformers with diffuse mechanics or regenerate behaviors of one pre-existing game, our work proposes an open-ended narrative experience with stable game mechanics enabled by an LLM-based game engine. 

\begin{figure}[t!]
    \centering
    \includegraphics[width=0.9\textwidth]{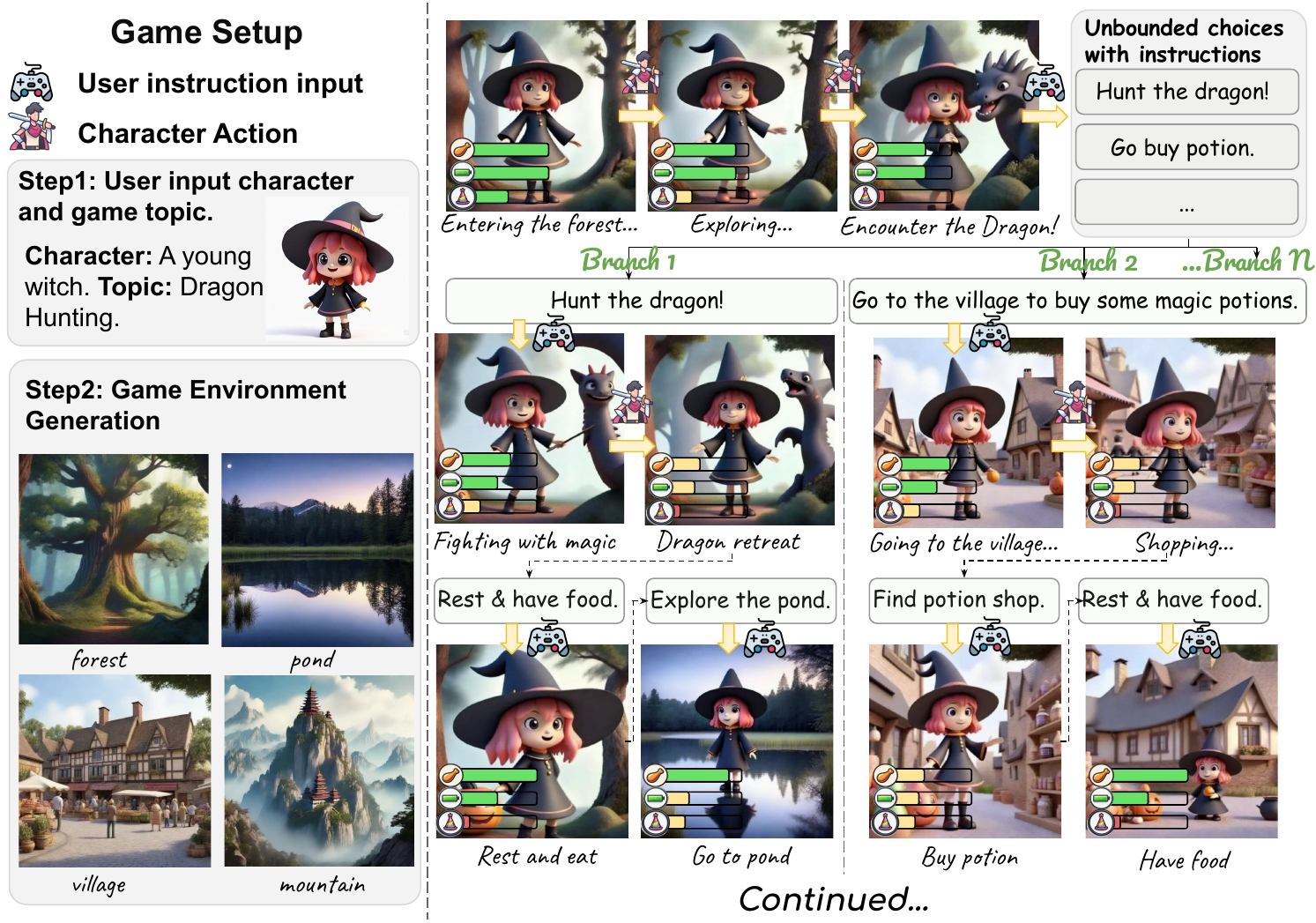}
    \caption{Example of \methodname{}. Based on an initial user input, \methodname{} sets up game simulation environments, and generates character actions in the environments. Users can interact with the character with natural language instructions, exploring the game with unlimited options.
    }
   \label{fig:teaser2}
   \vspace{-8pt}
\end{figure}

\begin{figure}[t!]
    \centering
    \includegraphics[width=0.9\textwidth]{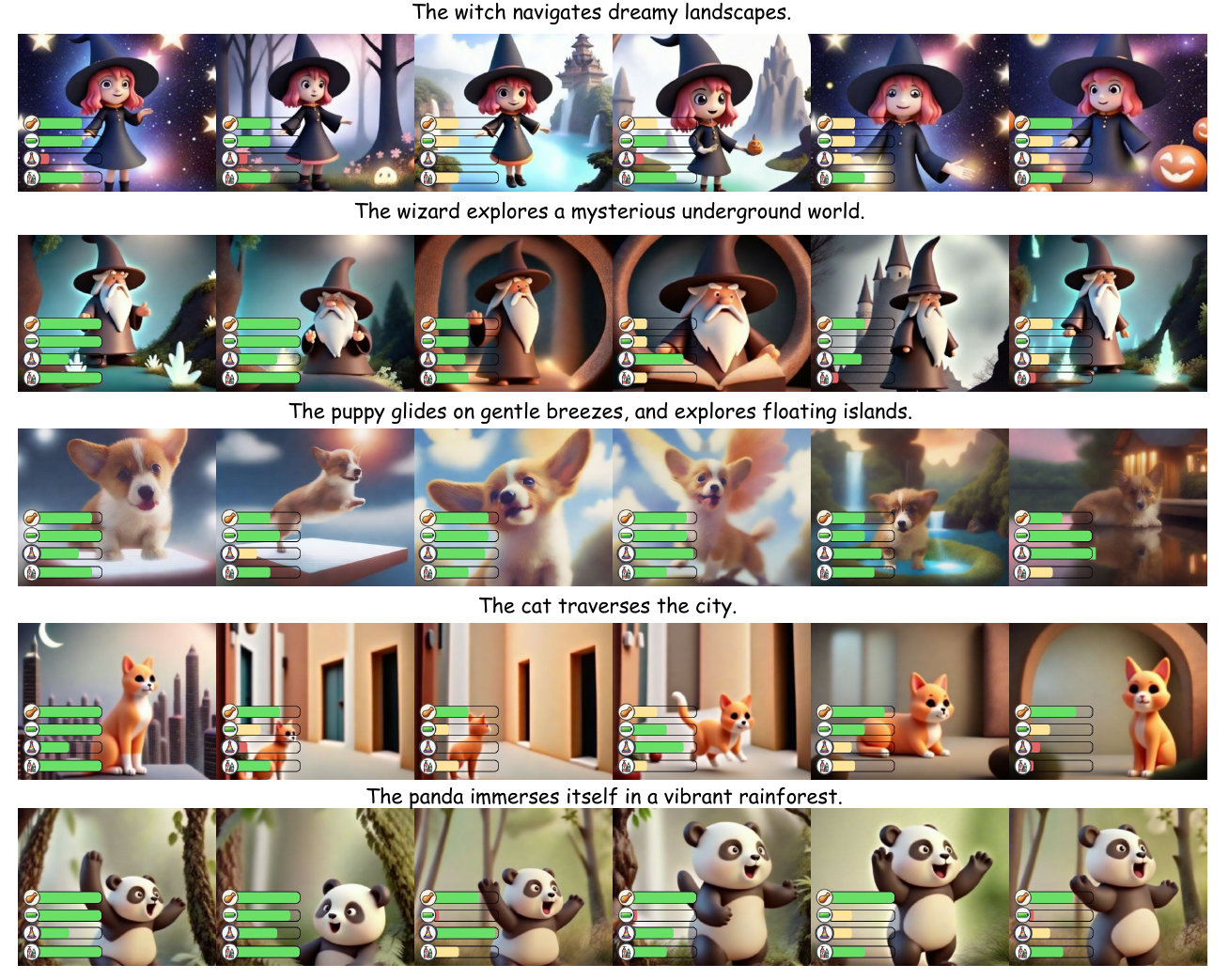}
    \caption{Generative game examples of \methodname{}. The user can insert a custom character into the game, engage with the character through natural language instructions, bring the character to different environments, and interact with it to maintain a healthy state under the games' mechanics.
    }
   \label{fig:universe}
   \vspace{-8pt}
\end{figure}

\methodname{} offers a gameplay loop centered around character simulation and open-ended interaction (Figure~\ref{fig:teaser2}). Players can insert their characters into the game, defining their appearance and personality. The game generates a world where these characters can explore environments, interact with objects, and engage in conversations. The game generates new scenarios, stories, and challenges based on the player's actions and choices, creating a personalized and infinite gaming experience. Some generative game examples are shown in Figure~\ref{fig:universe}.

Specifically, \methodname{} has the following capabilities: (1) \textbf{Character Personalization}: players can insert their characters into the game, defining their appearance and personality. (2) \textbf{Game Environment Generation}: \methodname{} generates a persistent world that the characters can explore and interact. (3) \textbf{Open-Ended Interaction}: Players can interact with the character using natural language instructions, and there are no pre-defined rules to constraint the interaction. (4) \textbf{Real-Time Generation}: we pay special attention to the speed of the game and achieve 5-10x speedups over a naive implementation, serving each new scene with a latency of about \textit{one second}.

Our approach introduces technical innovations in both LLM and vision generation domains. On the language side, we developed a \textbf{LLM based game engine} capable of maintaining consistent game mechanics, generating coherent narratives, and producing contextual character responses in real-time. Our distilled specialized model is fine-tuned on data automatically generated with two collaborative strong LLM agents, without the need for human annotation in the loop. Our distilled LLM model handles the dynamic generation of game rules and scenarios, adapting to player input and game state. In visual generation, we introduce a new \textbf{regional IP-Adapter}, which allows for the consistent generation of characters and environments while maintaining visual coherence across multiple images. Specifically, our regional IP-Adapter conditions the image generation on the game environment and character appearance encoded modulated by a dynamic mask obtained from attention outputs in cross-attention layers. This is in order to mitigate the interference between the environment and character, in order to have both reliably appear in the scene. This approach enables real-time image generation that reflects the game state and player actions.

The contributions of this work are conceptual and technical. We introduce the notion of a generative infinite game, demonstrating its feasibility and potential impact on the future of interactive entertainment. We present a new paradigm for game design where the game logic and content are encapsulated within generative models. Our main technical contributions include the specialized distilled LLM for game logic and narrative generation and the regional IP-Adapter for consistent visual generation. We demonstrate the effectiveness of our regional IP-Adapter through both quantitative and qualitative evaluations, surpassing state-of-the-art in both character and environment consistency. Furthermore, we show that our distilled LLM performs comparably to a very large LLM while having interactive speed. These advancements enable the creation of \methodname{} and lay the groundwork for future research and development in the field of AI-driven interactive experiences.

\section{Related Work}

\paragraph{Controllable Text-to-Image Generation.} Controllable text-to-image generation becomes a key research direction in diffusion model applications, enabling diverse ways to guide the generation process. For instance, ControlNet~\citep{zhang2023adding} introduces conditioning mechanisms that utilize control signals such as depth maps, poses, edges, and segmentation maps to guide image generation. Other works focus on layout control using bounding boxes to control object placement within the generated images~\citep{li2023gligen, shin2022reco}. Beyond these control signals, another major area of research involves personalization, where the goal is to generate consistent characters~\citep{ruiz2023dreambooth, gal2022image, kumari2023multi, avrahami2024chosen} or consistent face identities~\citep{li2024photomaker, wang2024instantid, yan2023facestudio, ruiz2024hyper} across multiple generations. However, most existing approaches lack support for conditioning both characters and environments separately, often requiring predefined masks for generating characters~\citep{chen2024anydoor, lugmayr2022repaint, yang2023paint}, with the environment remaining identical to the input image. This limitation makes it difficult to seamlessly integrate characters into different environments while ensuring both consistency and alignment with the input prompts. 
IP-Adapter~\citep{ye2023ip} tackles this task by conditioning the generation on the environment and character images. However, IP-Adapter tends to over reconstruct the conditions, which causes interference between them. In this paper, we build our approach on IP-Adapter and propose an improved regional IP-Adapter with block drop, separating character and environment generation to enhance consistency.

\paragraph{Large Language Models in Image Generation.} Large language models have demonstrated strong in-context learning capability~\citep{brown2020language}, which enables them to solve diverse customized tasks based on human instructions and in-context examples. In the field of image generation, large language models have been employed for various tasks, such as image layout generator based on prompts~\citep{cheng2024autostudio, cheng2024theatergen}, interactive multi-turn image generation~\citep{zeqiang2023mini, huang2024dialoggen, gong2023talecrafter, wang2023autostory}, and interleaved text and image generator~\citep{team2024chameleon, zhou2024transfusion}. Unlike these applications, our work focuses on distilling a specialized LLM, serving as a game engine, responsible for generating game mechanics, narratives and character interactions.

\paragraph{Game Generation.} In the field of Procedural Content Generation (PCG), diverse approaches have been explored to create dynamic game content~\citep{summerville2018procedural}.
Early works employed concept maps~\citep{treanor2012game}, Markov Chains~\citep{snodgrass2014experiments}, Bayes Nets~\citep{guzdial2016game}, and LSTMs~\citep{summerville2016super} to simulate interactive game environments. 
Recent research has advanced to using Generative Adversarial Networks (GANs) for generating game levels or dynamic environments~\citep{volz2018evolving, kumaran2019generating, schubert2021toad, kim2020learning}, exploiting the use of diffusion models for game generation~\citep{zhou2024eyes}, utilizing LLMs to design and generate the game environments or mechanics~\citep{sudhakaran2024mariogpt, todd2023level, nasir2023practical, hu2024game, zala2024envgen, anjum2024ink, chung2024patchview, chung2024toyteller} and employing LLMs to generate items and narratives, as well as using diffusion models for scene generation in 1001 Nights~\citep{sun2023language}. 
One major feature of this body of work is that AI is often employed to help in game design, or to generate one component of the game, for example environments or interactions. Our work aims to fully generate all behaviors of the game, including all the game's graphics, characters, environments, and narrative, using generative models. 

\citet{bruce2024genie} explore synthesizing a completely new interactive game with a video diffusion model conditioned on previous game frames and user actions. This work achieves impressive initial results for this direction, at present it has limitations on the scope of the games being generated, with mostly 2D platformers being represented. This differs from our work since these are not infinite games. We also believe that generating compelling infinite games by only modeling pixels is a hard problem and hence we opt to use language models to generate all the open-ended game mechanics instead. \citet{valevski2024diffusion} train a diffusion model to be a game engine running Doom, and achieve consistent generation of game mechanics by only modeling pixels. Nevertheless, this work currently fits the diffusion model to a single finite game that already exists.

Besides generating the game content, there has been exploration in letting AI take different roles (e.g., competitor, designer, or teammate) in games~\citep{zhu2021player, gallotta2024large, pell1992metagame, agarwal2023controllable}, and a system with continuous creativity for automated game design for games that are played on a grid of up to 8x8 squares is built in ~\citet{cook2022puck}.
Alongside generating visual game environments, another line of research builds narrative text games with conditional language generation models to complete a knowledge graph~\citep{ammanabrolu2020bringing} or utilizes LLMs or other AI tools to generate the game~\citep{aitamago, aidungeon, wang2023bytesized32} or act in different roles in the game~\citep{zhu2023calypso, kreminski2020we, cui2023thespian,zhou2023dialogue,dambekodi2020playing}. Our work diverges from these by (1) tackling both generation of all game mechanics, graphics, characters, environments, and narrative of the game using generative models (2) introducing personalization into generative games, where a user can insert a custom character and personalize the game world and initial story arc, and (3) achieving interactive real-time refresh speed through our innovations in the vision and language domains.

\section{Method}
We introduce \methodname{}, an interactive generative infinite game powered by text-to-image generation models and large language models. \methodname{} offers: (1) \textbf{Custom Character Personalization}: users create unique characters with customizable appearances and personalities; (2) \textbf{Dynamic World Creation}: the system generates a persistent, interactive game world for exploration; (3) \textbf{Open-Ended Interaction and Gameplay}: players interact with their characters via natural language, with the game dynamically generating new scenarios and storylines based on player actions; and (4) \textbf{Generation in Interactive Speed}: the game runs with near real-time interactivity, achieving a refresh rate close to \textit{one second}. We detail the methods enabling these capabilities in this section.

\subsection{Personalization of Latent Consistency Models for Character Consistency} \label{sec:lcm}

A key feature of \methodname{} is its ability to serve a fully generative model-based game with real-time interaction. This is made possible through the use of latent consistency models (LCM)~\citep{luo2023latent} which allow for high-resolution image generation with as few as two diffusion steps. By utilizing LCMs, \methodname{} achieves real-time text-to-image (T2I) generation, critical for delivering an interactive gaming experience with a refresh rate close to \textit{one second}.

To support the use of custom characters in the game, we incorporate DreamBooth~\citep{ruiz2023dreambooth} into the T2I model. Given a set of character images, we fine-tune the diffusion model using LoRA modules~\citep{hu2021lora}. During fine-tuning, we append a unique identifier, ``[V]'', which has a weak prior in the model, to denote the subject. DreamBooth personalization is performed on the base diffusion model, and we merge the subject-specific LoRA with the LCM LoRA trained for few-step diffusion. This simple arithmetic LoRA merging works surprisingly well in maintaining both inference speed and subject preservation. Other alternatives for personalization exist, and many do not need setup time, yet we find that they often fail in strongly preserving the character's features, which is a critical component in order to have a satisfactory experience in this type of game.

\begin{figure}[t!]
    \centering
    \includegraphics[width=0.85\textwidth]{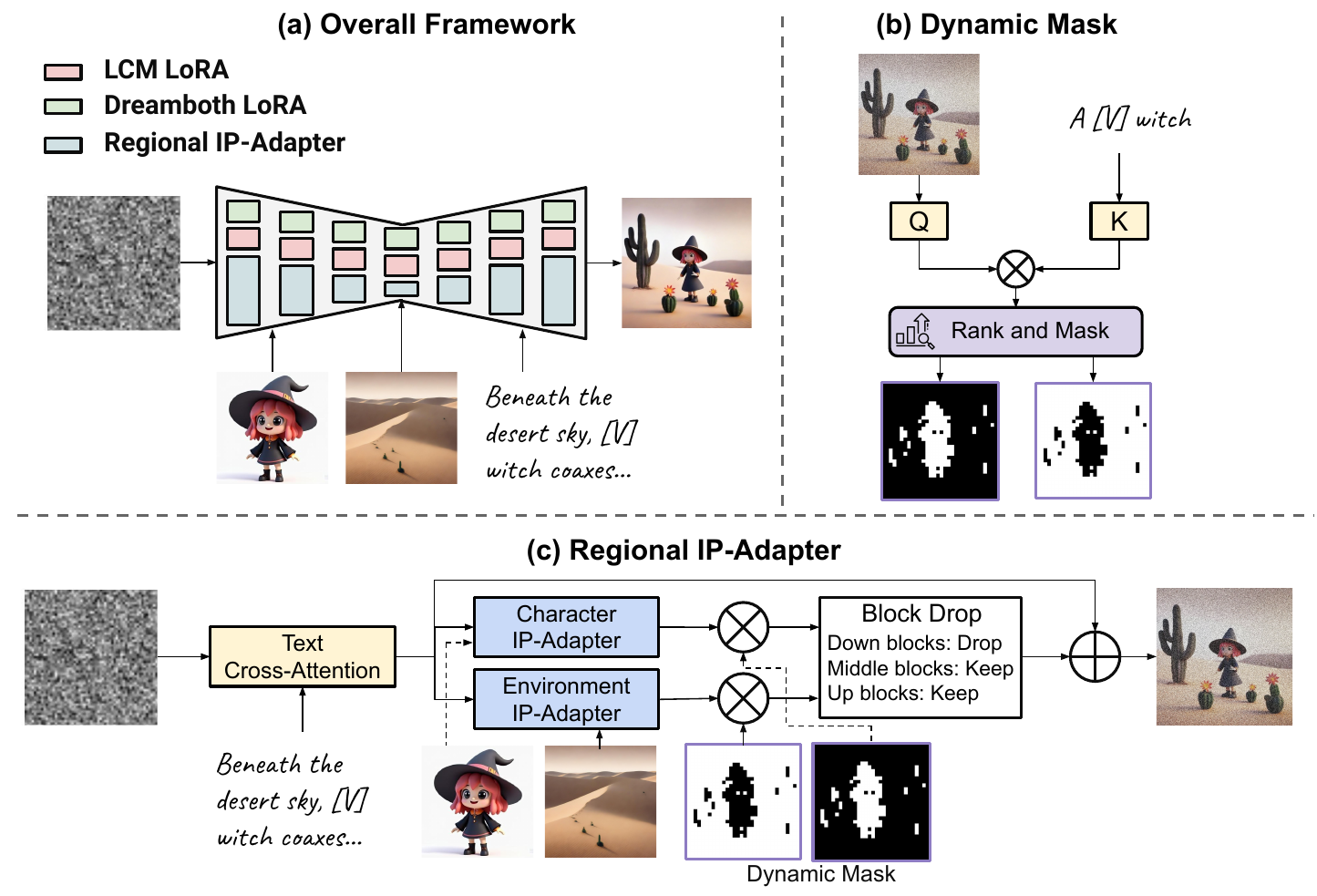}
    \caption{
    (a) Our overall image generation method. We achieve real-time image generation with LCM LoRA, maintain character consistency with DreamBooth LoRAs, and introduce a regional IP-Adapter (shown in (c)) for improved environment and character consistency. (b) Our proposed dynamic mask genreation separating the environment and character conditioning, preventing interference between the two.
    }
   \label{fig:model}
\end{figure}

\subsection{Regional IP-Adapter with Block Drop for Environment Consistency}

Another key feature of \methodname{} is generating the character in pre-defined environments performing different actions based on user instructions. Thus, maintaining both character and environment consistency is essential. While character consistency is handled as discussed in Sec.~\ref{sec:lcm}, two additional challenges arise: ensuring the environment consistency across different generations and accurately placing the character within the environment, without losing alignment with text prompts. We find that existing method fails to consistently perform well for all requirements in interactive speed. As one of our main technical contributions we propose a novel \textbf{regional IP-Adapter} in order to consistently implant a character in pre-defined environments following text prompt.

\subsubsection{Regional IP-Adapter}
We propose an improved version of IP-Adapters~\citep{ye2023ip} that enables dual-conditioning on both subjects and environments, allowing for the generation of a pre-defined character in a user-specified environment. Unlike the original IP-Adapters, which focus on single-image conditioning, our approach introduces dual-conditioning and dynamic regional injection mechanism to represent both concepts simultaneously in the generated images.

Let us start with an example. As shown in Figure~\ref{fig:model}, given the text prompt \textit{``Beneath the desert sky, [V] witch coaxes the cacti to blossom with vibrant, glowing flowers''} and the desert environment image, the model needs to know that the character should be generated beside cacti and within the desert environment. This requires the model to correctly (1) preserve the environment (2) preserve the character (3) follow the prompt. Utilizing IP-Adapter to encode the environment greatly harms both (2) and (3) (Figure~\ref{fig:example}). Our regional IP-Adapter solves this by implementing a novel attention separation mechanism for generating the two elements. Specifically, we introduce a dynamic mask-based approach that leverages cross-attentions between the character text embedding and the hidden states at each layer of the model. As shown in Figure~\ref{fig:model}, our approach applies the adapter to the regions corresponding to the environment and character separately, preventing the environment conditioning to interfere with the character's appearance, and vice versa. At each cross-attention layer, we calculate the attention map between a pre-defined character text embedding $K_c$ and the output hidden states from the text cross-attention layer $O_t$:

\begin{equation}
A_c = \frac{W_qO_t * W_kK_c^T}{\sqrt{d}}
\end{equation}

where $W_q$ and $W_k$ are projection weights adopted from the text cross-attention layers. The dynamic mask we use for the regional IP-Adapter is defined as:

\begin{equation}
M_c = 
\left\{
\begin{array}{ll}
    1 & A_c \leq  threshold \\
    0 & A_c > threshold \\
\end{array}
\right.
\end{equation}

We rank the attention scores $A_c$ and set the $threshold$ at top $r\%$, dynamically updating the mask at each attention layer throughout the diffusion process.
The output of the cross-attention blocks is calculated as:

\begin{equation}
O = O_t + \alpha_{e} M_c * O_e + \alpha_{c} (1 - M_c) * O_c
\end{equation}

where $O_e$ and $O_c$ represent the outputs of the IP-Adapter image cross-attention layers of environment and character, and $\alpha_{e}$ and $\alpha_{c}$ are the IP-Adapter scales to adjust the strength of the environment and character conditioning respectively. Our regional IP-Adapter with dynamic masks allows the model to generate the character without the interference from the environment conditioning, greatly enhancing character consistency preservation (Figure~\ref{fig:example}). Besides, it preserves environment consistency by generating the background around the character conditioning on the environment image, while maintaining the implicit layout information in the hidden states from the text cross-attention layers $O_t$, ensuring the character is placed accurately following the text prompt.

\begin{wrapfigure}{r}{0.6\textwidth}
    \begin{center}
    \vspace{-5pt}
    \includegraphics[width=.6\textwidth]{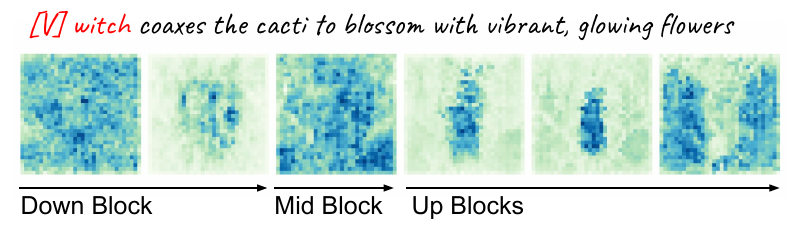}
    \end{center}
    \vspace{-10pt}
    \caption{
    Attention map between character embedding and hidden states in cross-attention layers in different blocks. The character embedding we use is ``A [V] witch''.
    }
    \vspace{-10pt}
   \label{fig:attn}
\end{wrapfigure}

\subsubsection{Block-Wise Drop on Environment Conditioning}

For our regional IP-Adapter, we use a dynamic mask from the cross-attention between character text and hidden states. This mask's quality is key for separating character and environment generation. Figure~\ref{fig:attn} shows attention maps between character embeddings and hidden states in cross-attention layers of down sample blocks. We observe that the attention doesn't focus on the character but spreads across the full image for these blocks. We take this as a strong indication that the diffusion model doesn't separate character and environment generation in these layers and instead focuses on overall image structure based on text prompts. This aligns with \citet{wang2024instantstyle}'s finding that down sample blocks capture spatial layouts more, while up sample blocks capture style. We drop the regional IP-Adapter in down sample blocks, using it only in mid and up sample blocks. This allows for better spatial layout generation between character and environment.
Additionally, we find that adding dynamic regional IP-Adapters in up sample blocks more strongly aligns the generated background with the conditioning environment by extracting relevant semantic information while preserving character-specific details like appearance and poses.

\subsection{Language Model Game Engine with Open Ended Interactions and Integrated Game Mechanics} 
\label{sec:lang}
\methodname{} simulates character actions in pre-defined environments with images generated based on scene text descriptions while monitoring the character's state and providing the user with natural language interaction with the character and game environment. For example, the user can take the character to different environments, interact with the character (e.g. \textit{``I pet the character on its head''}), or essentially take any open ended action e.g. \textit{``I pet the character and then take them for a rocket ride at the space station.''}. Since the game is ultimately built on a language model, these expansive capabilities present several challenges: (1) \textit{Environment Binding}: the model needs to place the character in the correct environment based on the natural language instructions. (2) \textit{Coherent Story Generation}: the model generates coherent narrative descriptions and character responses that align with user-specified character traits. (3) \textit{Game Mechanics Generation}: the model needs to monitor the state of the character (i.e., hunger, energy, fun, hygiene), and update them based on user interactions and story events. (4) \textit{Prompt Rewriting}: the model needs to rewrite the narratives for the diffusion model (i.e., append special token ``[V]'' for the character, align environment descriptions to the pre-generated environments for better environment consistency).

Surprisingly, we find that a very large language model (e.g., GPT-4, GPT-4o) with detailed instructions and using in-context learning~\citep{brown2020language} can exhibit these capabilities. Nevertheless, using such large models as game engines is not directly feasible due to the large latency (e.g., 5 seconds for a 7B model to give one response). Given this, we propose to distill these capabilities from a very large model into a smaller model based on Gemma-2B~\citep{team2024gemma} for game logic and narrative generation that supports real-time interaction. In this section we propose two key technical contributions (1) a design for a character life simulation game using two very large language models that control for world modeling and user interaction respectively (2) a framework for distilling this knowledge into one smaller language model that is fast enough to achieve interactive speeds.

\subsubsection{Character Life Simulation with Multi-LLM Collaboration} 

We build a character life simulation game with two LLMs agents. One agent serves as the world simulation model, responsible for setting up game environments, generating narratives and image descriptions, tracking character states and simulating the character's behavior. The second agent functions as a user model, simulating the player's interactions with the world simulation model. It has three types of interactions: continuing the story within the current environment, moving the character into different environments, or interacting with the character to maintain the healthy state of the character. In each interaction category, the user has the option to provide personality details of the character or guide the character actions that, in turn, guide the simulator’s narrative generation. This interaction between the world simulation LLM and the user LLM allows for a dynamic character life simulation game with virtually unlimited interaction possibilities and narrative paths.

\begin{figure}[t!]
    \centering
    \includegraphics[width=0.8\textwidth]{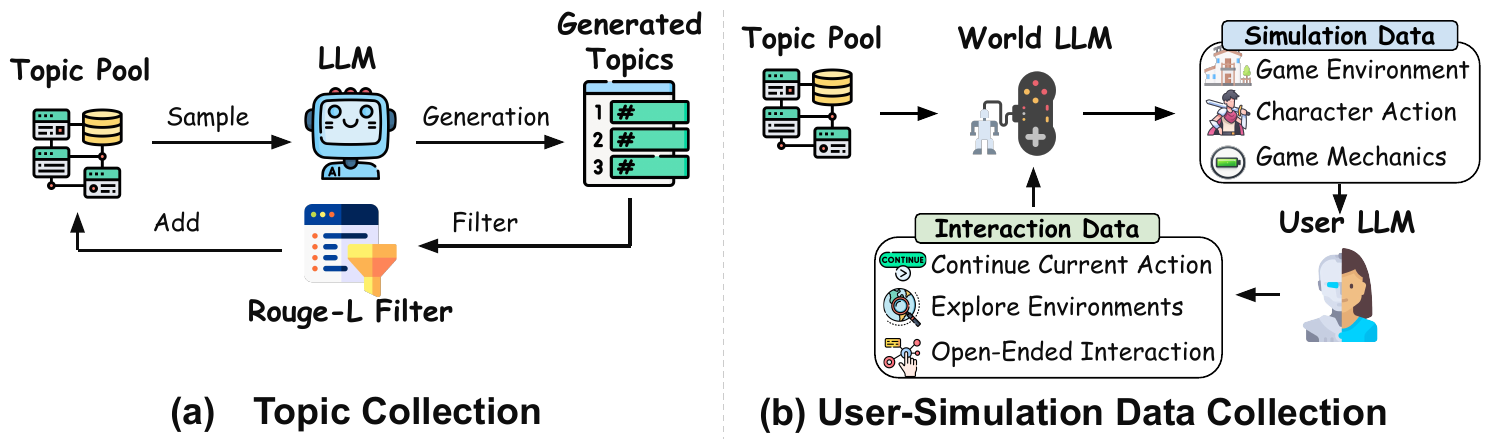}
    \vspace{-5pt}
    \caption{ 
    Overview of our user-simulation data collection process for LLM distillation. (a) We begin by collecting diverse topic and character data, filtered using ROUGE-L for diversity. (b) The World LLM and User LLM interact to generate user-simulation data through multi-round exchanges.
    }
    \vspace{-10pt}
   \label{fig:llm}
\end{figure}

\subsubsection{Framework for Small LLM Distillation} 

We propose a framework for distilling the capabilities of larger LLMs into the smaller, more efficient model using synthetic data generated by the multiple stronger LLMs. Our framework contains two steps: (1) automated data collection (Figure~\ref{fig:llm}) and (2) small model distillation.

\paragraph{Automated Data Collection}
Our goal is to build a general-purpose character life simulator capable of understanding a wide range of character traits and generating games across diverse topics. To achieve this, the first step is to gather a diverse dataset of topics and characters. We prompt a large LLM to generate pairs of topics and corresponding main characters. To ensure data diversity, we retain only the generated pairs whose ROUGE-L similarity to existing data is below 0.7, following~\citep{wang2022self}, which demonstrated the importance of diverse data for enhancing an LLM’s ability to follow instructions. This process results in 5,000 unique topic-character pairs, which serve as the basis for user-simulator interaction data. 

In the second step, we collect multi-round interaction data between the world simulation LLM and the user LLM. The process begins with the world simulation LLM setting up the game environment and initiating a character action based on a randomly sampled topic-character pair from the dataset. The user LLM is then prompted to provide interaction inputs, while the world simulation LLM generates updated character actions, states, and responses. This iterative process continues for five interaction rounds per session, resulting in a total of 5,000 user-simulator interaction examples. All the prompt templates are in Appendix. 

\paragraph{Distillation}
Once the interaction data has been collected, we fine-tune the smaller Gemma-2B model using the 5,000 synthetic user-simulator interaction samples. During supervised fine-tuning, we mask out the loss on user input data, focusing the optimization on learning the world simulation model’s behavior based on multi-round interaction history and current user input. This approach enables Gemma-2B to replicate the capabilities of larger LLMs as a game engine while supporting real-time interaction. Our distilled Gemma-2B demonstrates performance comparable to GPT-4o, effectively following user input and supporting unbounded interactions.

\section{Experimental Setup}

\subsection{Evaluation Benchmarks} 

\paragraph{Evaluation of Image Generations}  To evaluate our image generation approach, we collect an evaluation dataset consisting of 5,000 (character image, environment description, text prompt) triplets with GPT4o~\citep{gpt}. It includes 5 characters (dog, cat, panda, witch, and wizard), 100 diverse environments, and 1,000 text prompts (10 per environment). We evaluate the image generation performance under three criteria: environment consistency, character consistency and semantic alignment with the text prompt. To measure similarity between images, we employ CLIP-I~\citep{radford2021learning}, DINO~\citep{caron2021emerging}, and DreamSim~\citep{fu2023dreamsim}. We denote the similarity between environment reference image and the generated images as CLIP-I$^{E}$, DINO$^{E}$, DreamSim$^{E}$, and the similarity between the character reference image and the generated images as CLIP-I$^{C}$, DINO$^{C}$, DreamSim$^{C}$. Additionally, we use CLIP-T~\citep{radford2021learning} to evaluate the semantic alignment with the text prompt. Given that \methodname{} is a character life simulation game, ensuring the presence of the character in the image is important. Therefore, we further utilize Grounding-DINO~\citep{liu2023grounding} to detect the presence of the character in the generated images. We set similarity scores to 0 and distance scores to 1 if there is no character in the generated image.

\paragraph{Evaluation of LLM Generations} We collect an additional evaluation dataset with 100 user-simulator interaction samples using the pipeline in Sec.
~\ref{sec:lang}. Each user-simulator interaction sample contains five rounds of interaction between the user and the world model. We use GPT-4~\citep{gpt} as a judge, scoring the response between two models (baseline model vs. our model) in overall score, and then in four aspects: accuracy of character state update, environment relevance, story coherence, and user input instruction following. Scores range from 0 to 10. 

\subsection{Implementation Details} \label{sec:implementation}

Our image generator is built on SDXL~\citep{podell2023sdxl}. 
We train a DreamBooth LoRA of rank 16 with batch size 1 and a constant learning rate 1e-4 for 500 steps on a single A100, which takes approximately 30 mins. The special token we append before the character is ``sks''. During inference, we merge the LCM-LoRA with DreamBooth LoRA with scale 1.0 for each. We use IP-Adapter-plus-sdxl-vit-h for encoding the environment, and IP-Adapter-plus-face-sdxl-vit-h for encoding the character. The dynamic mask ratio $r\%$ in set to be 60\%. 

Our LLM is built on Gemma-2B~\citep{team2024gemma}. We distill the LLM using 5,000 user-simulator interaction samples collected from GPT-4o. We train the LLM for 6,500 steps, with batch size 8, distributed across 4 A100s, and learning rate 1e-4. The learning rate scheduler is set to be cosine annealing~\citep{loshchilov2016sgdr}, and the warmup steps ratio is 0.03. In evaluation, we use an LLM to generate responses with sampling. The sampling hyperparameters are set to be default.

\section{Results and Analysis}

\begin{table}[t]
  \caption{Comparison of \methodname{} and other methods for maintaining environment consistency and character consistency. 
  \methodname{} achieves the best performance in maintaining consistency, while maintaining comparable semantic alignment with the text prompt. Best scores are in \textbf{bold}.
  }
  \vspace{5pt}
  \label{table:main_quantitative}
  \resizebox{\columnwidth}{!}{
  \centering
  \begin{tabular}{l  ccc  ccc  c}
    \toprule 
    \multirow{2}{*}{\textbf{Methods}} & \multicolumn{3}{c}{\textbf{Environment Consistency}} & \multicolumn{3}{c}{\textbf{Character Consistency}} & \multicolumn{1}{c}{\textbf{Semantic Alignment}} \\ 
   \cmidrule(lr){2-4} \cmidrule(lr){5-7} \cmidrule(lr){8-8}
  &  CLIP-I$^{E}$ $\uparrow$ & DINO$^{E}$ $\uparrow$ & DreamSim$^{E}$ $\downarrow$ & CLIP-I$^{C}$ $\uparrow$ & DINO$^{C}$ $\uparrow$ & DreamSim$^{C}$ $\downarrow$ & CLIP-T $\uparrow$ \\
    \midrule
    IP-Adapter~\citep{ye2023ip} & 0.470 & 0.381 & 0.595 & 0.366 & 0.139 & 0.832 & 0.168  \\ 
    IP-Adapter-Instruct~\citep{rowles2024ipadapter} & 0.334 & 0.151 & 0.832 & 0.246 & 0.124 & 0.872 & 0.098 \\ 
    StoryDiffusion~\citep{zhou2024storydiffusion} & 0.528 & 0.257 & 0.733 & 0.629 & 0.464 & 0.545 & \textbf{0.242} \\
    Ours & \textbf{0.563} & \textbf{0.322} & \textbf{0.675} &\textbf{0.676} & \textbf{0.470} & \textbf{0.488} & \textbf{0.242} \\ 
    \bottomrule
  \end{tabular}
  }
  \vspace{-10pt}
\end{table}

\begin{figure}[t!]
    \centering
    \includegraphics[width=0.85\textwidth]{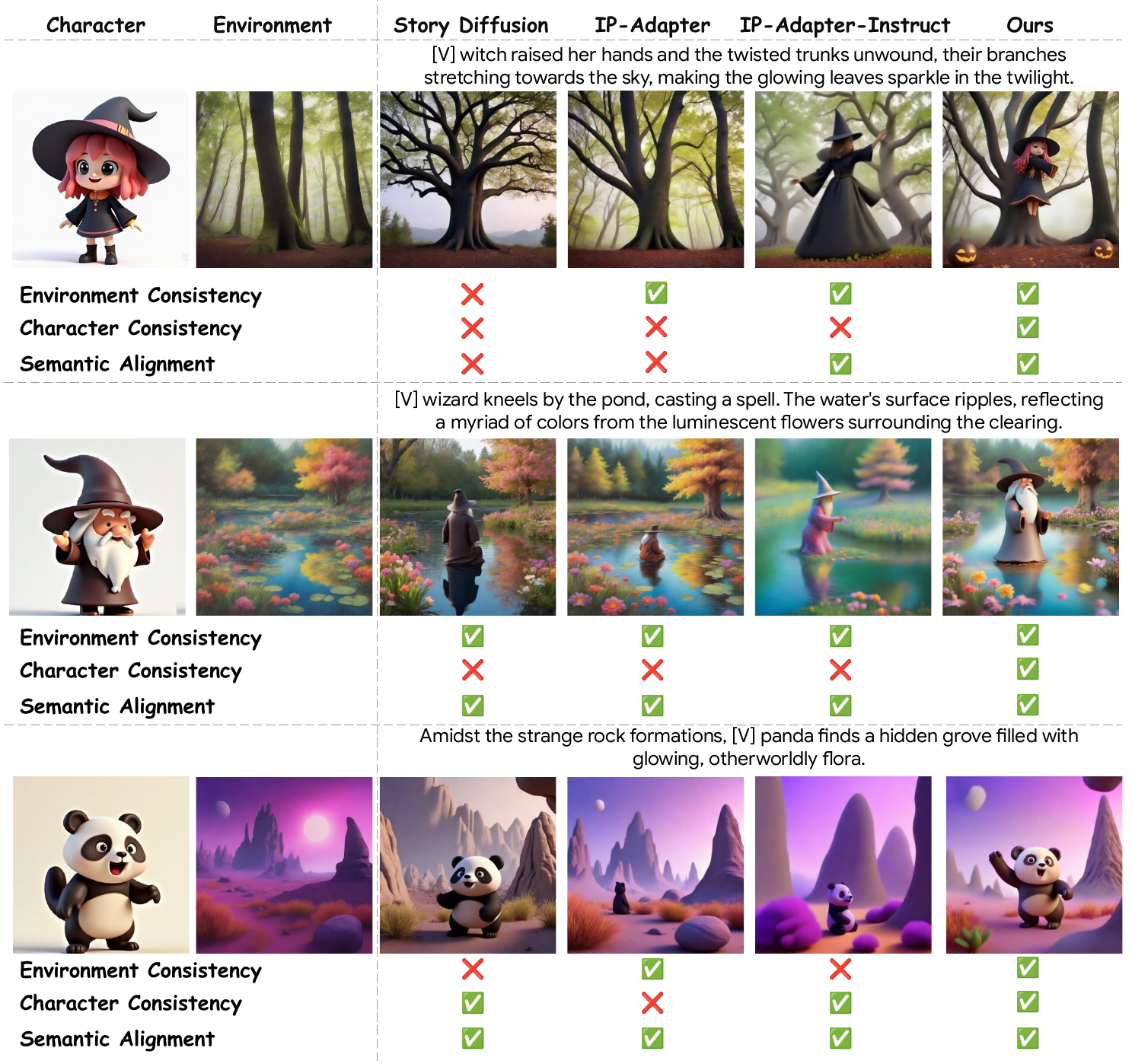}
    \vspace{-5pt}
    \caption{Comparison with other approaches for generating environment and character consistent images based on text prompts. We observe that our method strongly outperfoms related work.
    }
    \vspace{-10pt}
   \label{fig:comp}
\end{figure}

\begin{table}[t]
  \caption{Ablations of the effectiveness of dynamic regional IP-Adapter and block drop in our consistent image generation approach. 
  }
  \vspace{5pt}
  \label{table:ablation}
  \resizebox{0.95\columnwidth}{!}{
  \centering
  \begin{tabular}{l ccc ccc  ccc  c}
    \toprule 
   \multirow{2}{*}{\textbf{No.}} &  \textbf{Block} & \textbf{Regional} & \multirow{2}{*}{\textbf{Scale}}  & \multicolumn{3}{c}{\textbf{Environment Consistency}} & \multicolumn{3}{c}{\textbf{Character Consistency}} & \multicolumn{1}{c}{\textbf{Alignment}} \\ 
   \cmidrule(lr){5-7} \cmidrule(lr){8-10} \cmidrule(lr){11-11}
  & \textbf{Drop} & \textbf{IP-Adapter} & & CLIP-I$^{E}$ $\uparrow$ & DINO$^{E}$ $\uparrow$ & DreamSim$^{E}$ $\downarrow$ & CLIP-I$^{C}$ $\uparrow$ & DINO$^{C}$ $\uparrow$ & DreamSim$^{C}$ $\downarrow$ & CLIP-T $\uparrow$ \\
    \midrule
    1. & \xmark & \xmark & 1.0 & 0.123 & 0.111 & 0.885 & 0.073 & 0.024 & 0.973 & 0.034 \\ 
    2. & \cmark & \xmark & 1.0 & 0.414 & \textbf{0.331} & \textbf{0.647} & 0.337 & 0.147 & 0.832 & 0.149 \\ 
    3. & \cmark & \cmark & 1.0 & \textbf{0.563} & 0.322 & 0.675 &\textbf{0.676} & \textbf{0.470} & \textbf{0.488} & \textbf{0.242} \\
    \midrule 
    4. & \xmark & \xmark & 0.5 & 0.470 & 0.381 & 0.595 & 0.366 & 0.139 & 0.832 & 0.168 \\
    5. & \cmark & \xmark & 0.5 & \textbf{0.577} & \textbf{0.332} & \textbf{0.640} & 0.627 & 0.374 & 0.575 & \textbf{0.252} \\ 
    6. & \cmark & \cmark & 0.5 & 0.549 & 0.263 & 0.726 & \textbf{0.705} & \textbf{0.514} & \textbf{0.450} & 0.246 \\ 

    \bottomrule
  \end{tabular}
  }
  \vspace{-10pt}
\end{table}

\subsection{Comparison with Different Approaches for Maintaining Environment Consistency and Character Consistency}

\textbf{Quantitative Results} We compare our regional IP-Adapter with block drop with previous approaches in maintaining environment consistency and character consistency. For all the approaches, we merge the character LoRA and LCM LoRA with the model to support fast inference and improve character consistency and provide an apples-to-apples comparison. 
As shown in Table~\ref{table:main_quantitative}, our approach consistently outperforms previous approach in maintaining environment consistency and character consistency, while achieving comparable performance in maintaining semantic alignment. Specifically, our approach significantly overtakes StoryDiffusion~\citep{zhou2024storydiffusion} by 0.047 in CLIP-I$^C$, and 0.057 in DreamSim$^C$ for character consistency, and 0.035 in CLIP-I$^E$, 0.065 in DINO$^E$, and 0.058 in DreamSim$^E$ for environment consistency, demonstrating the effectiveness of our approach. Besides, our approach also achieves comparable performance in maintaining semantic alignment, suggesting strong text following capabilities.

\textbf{Qualitative Results}
We present a qualitative comparison with other approaches in Figure~\ref{fig:comp}. Our regional IP-Adapter with block drop consistently generates images with high character consistency, whereas other methods may fail to include the character or generate characters with inconsistent appearances (see Example 1 \& 2). Furthermore, we show that our approach balances environment consistency and character consistency well, while other approaches might generate environments that differ from the condition environment (e.g., StoryDiffusion in Example 1 \& 3).

\begin{figure}[t!]
    \centering
    \includegraphics[width=0.85\textwidth]{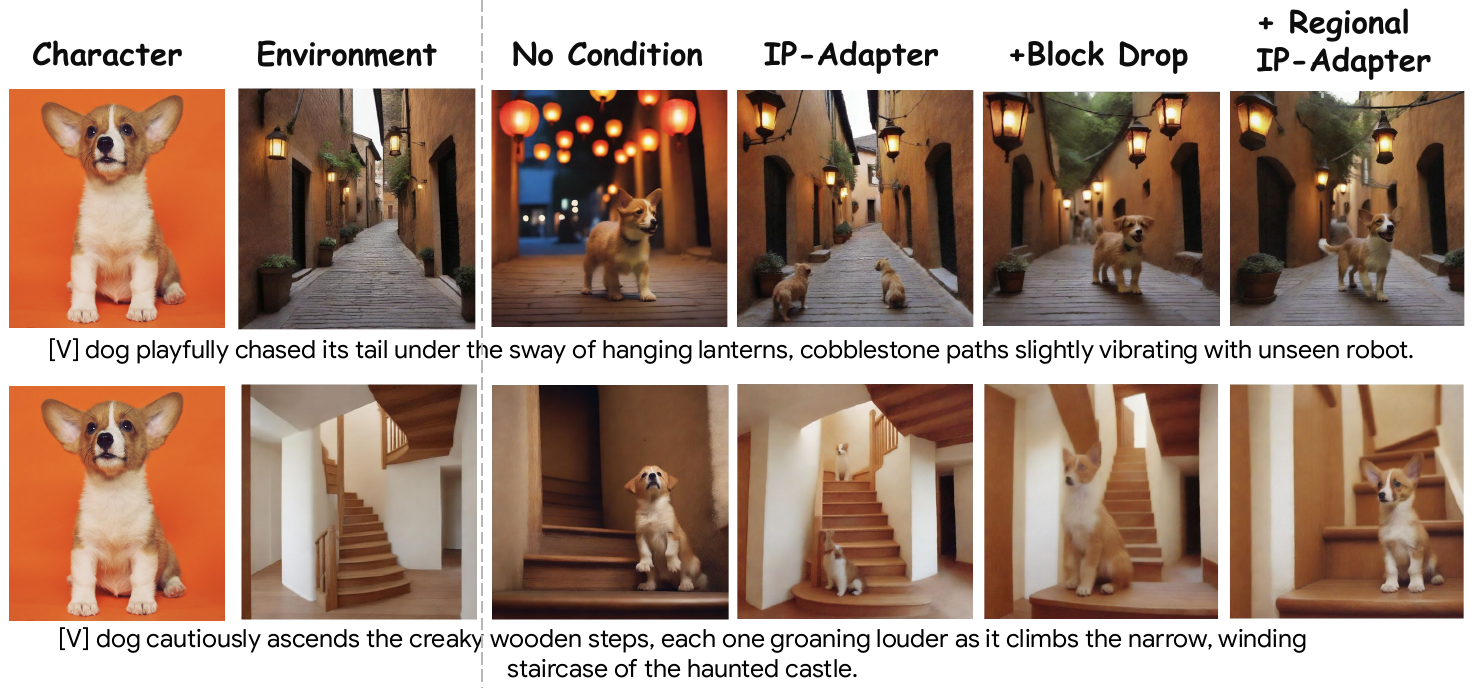}
    \vspace{-5pt}
    \caption{Comparison between our regional IP-Adapter approach and baseline approaches.
    }
    \vspace{-10pt}
   \label{fig:example}
\end{figure}

\subsection{Effectiveness of Dynamic Regional IP-Adapter with Block Drop}

\textbf{Quantitative Results} We demonstrate that our regional IP-Adapter with block drop is essential for placing the character in the environments following the text prompt, while maintaining both environment and character consistency with ablation studies. As shown in Table~\ref{table:ablation}, adding block drop improves both environment and character consistency compared with multi-IP-Adapter (No. 2. vs. No. 1.), with an increase of 0.291 in CLIP-I$^E$ and 0.264 in CLIP-I$^C$, alongside better alignment between the text prompt and the generated image. Furthermore, our regional IP-Adapter enhances character consistency and text alignment while maintaining comparable performance in environment consistency (No. 3 vs. No. 2). We also explore the effect of the environment IP-Adapter scale, and our findings indicate that using a smaller scale (e.g., 0.5) generally improves character consistency, though it slightly compromises environment consistency (No. 6 vs. No. 3).

\textbf{Qualitative Results} As shown in Figure~\ref{fig:example}, conditioning on the environment using IP-Adapter achieves good environment reconstruction, but the character consistency is influenced by the environment style. Introducing block drop improves adherence to the text prompt, resulting in images with the correct spatial layout for both the character and the environment. However, the character's appearance remains influenced by the surrounding environment. By incorporating our proposed regional injection mechanism with our proposed dynamic mask scheme, the generated images achieve strong character consistency while maintaining effective conditioning on the environment.

\begin{table}[t]
\caption{Comparison of \methodname{} and different LLMs on serving as game engines for open ended interactions and integrated game mechanics. We use GPT-4 to provide pairwise scores between our model and other LLMs.}
\vspace{5pt}
\centering
\resizebox{0.8\columnwidth}{!}{
\begin{tabular}{c cc cc cc cc cc }
    \toprule
   \multirow{2}{*}{\textbf{Model}} & \multicolumn{2}{c}{\multirow{2}{*}{\textbf{Overall}}} & \multicolumn{2}{c}{\textbf{State}} & \multicolumn{2}{c}{\textbf{Environment}} & \multicolumn{2}{c}{\textbf{Story}} & \multicolumn{2}{c}{\textbf{Instruction}}                 \\ 
   & & & \multicolumn{2}{c}{\textbf{Update}} & \multicolumn{2}{c}{\textbf{Relevance}} & \multicolumn{2}{c}{\textbf{Coherence}} & \multicolumn{2}{c}{\textbf{Following}} \\ 
   \cmidrule(lr){2-3}  \cmidrule(lr){4-5} \cmidrule(lr){6-7} \cmidrule(lr){8-9}  \cmidrule(lr){10-11}
   & Base & Ours & Base & Ours & Base & Ours & Base & Ours & Base & Ours \\ 
    \midrule
    Gemma-2B~\citep{team2024gemma} & 6.22 & \textbf{7.44} & 5.60 & \textbf{7.47} & 6.12 & \textbf{7.94} & 6.34 & \textbf{7.57} & 6.43 & \textbf{7.67} \\ 
    Gemma-7B~\citep{team2024gemma} & 6.80 & \textbf{7.39} & 6.29 & \textbf{7.43} & 7.07 & \textbf{7.91} & 6.90 & \textbf{7.48} & 6.89 & \textbf{7.53} \\ 
    Llama3.2-3B~\citep{llama} & 7.21 & \textbf{7.50} & 6.86 & \textbf{7.38} & 7.63 & \textbf{7.93} & 7.36 & \textbf{7.56} & 7.31 & \textbf{7.67} \\ 
    Ours-1k & 7.65 & \textbf{7.82} & 7.50 & \textbf{7.74} & 8.10 & \textbf{8.19} & 7.78 & \textbf{7.93} & 7.82 & \textbf{7.97} \\
    GPT-4o~\citep{gpt} & \textbf{7.76} & 7.68 & \textbf{7.69} & 7.66 & \textbf{8.20} & 8.10 & \textbf{7.95} & 7.82 & \textbf{7.85} & 7.82 \\ 
    \bottomrule
  \end{tabular}
  }
  \vspace{-10pt}
\label{tab:lang}
\end{table}

\subsection{Effectiveness of Distilling Specialized Large Language Model}

We show that our diverse user-simulator interaction data effectively distills Gemma-2B into a capable game engine. As shown in Table~\ref{tab:lang}, zero-shot inference with small LLMs (i.e., Gemma-2B, Llama3.2-3B), or a slightly larger LLM (i.e., Gemma-7B) results in lower performance compared to ours, highlighting the importance of distillation from a stronger LLM for game world and character action simulation. Furthermore, we show that our model achieves performance comparable to GPT-4o, validating the effectiveness of our approach. We also investigate the impact of distillation data size on performance by comparing a Gemma-2B model distilled with 1K data and 5K data. Results show that using a larger dataset consistently improves performance across all aspects, highlighting the potential for further enhancements with more data to fully match GPT-4o's performance. 

\section{Conclusion}

We introduce \methodname{}, an interactive generative infinite game based on generative models. \methodname{} is built on two main components, a specialized, distilled LLM for real-time interaction, and a fast diffusion model with our proposed regional IP-Adapter for consistent generation across multiple scenes. We show that our proposed approach allows for an interactive game subsumed in generative models, with consistent characters, environments and story and an expansive gameplay characteristic of an infinite game.

\section*{Acknowledgments}
We thank Shiran Zada, Peyman Millanfar, Shlomi Fruchter, Michael Goin and Matthew Guzdial for the thoughtful feedback and discussion.

\bibliography{iclr2025_conference}
\bibliographystyle{iclr2025_conference}

\newpage
\appendix
\section*{Appendix}

In this appendix, we present the following:
\begin{itemize}
    \item Prompts we use for user-simulation data collection in Sec.~\ref{sec:data_template}.
    \item Evaluation prompts we use for querying GPT-4 as a judge for LLM evaluation in Sec.~\ref{sec:llm_evaluation}.
    \item Reproducibility statement in Sec.~\ref{sec:reproduce}.
\end{itemize}

\section{Prompt for synthetic user-simulator interaction data collection} \label{sec:data_template}

\begin{figure}[h]
    \centering
    \includegraphics[width=\textwidth]{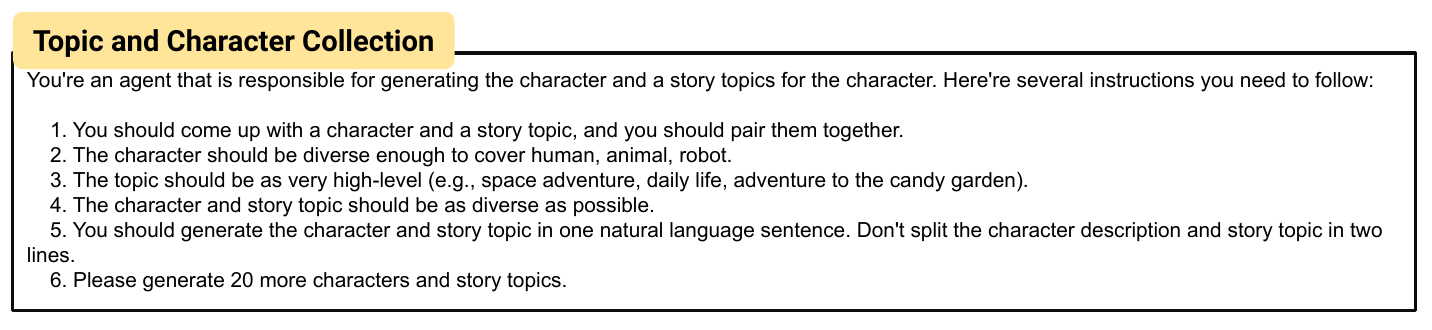}
    \caption{Prompt template used to collect diverse topic and character data.
    }
   \label{fig:topic_prompt}
\end{figure}

\begin{figure}[t!]
    \centering
    \includegraphics[width=\textwidth]{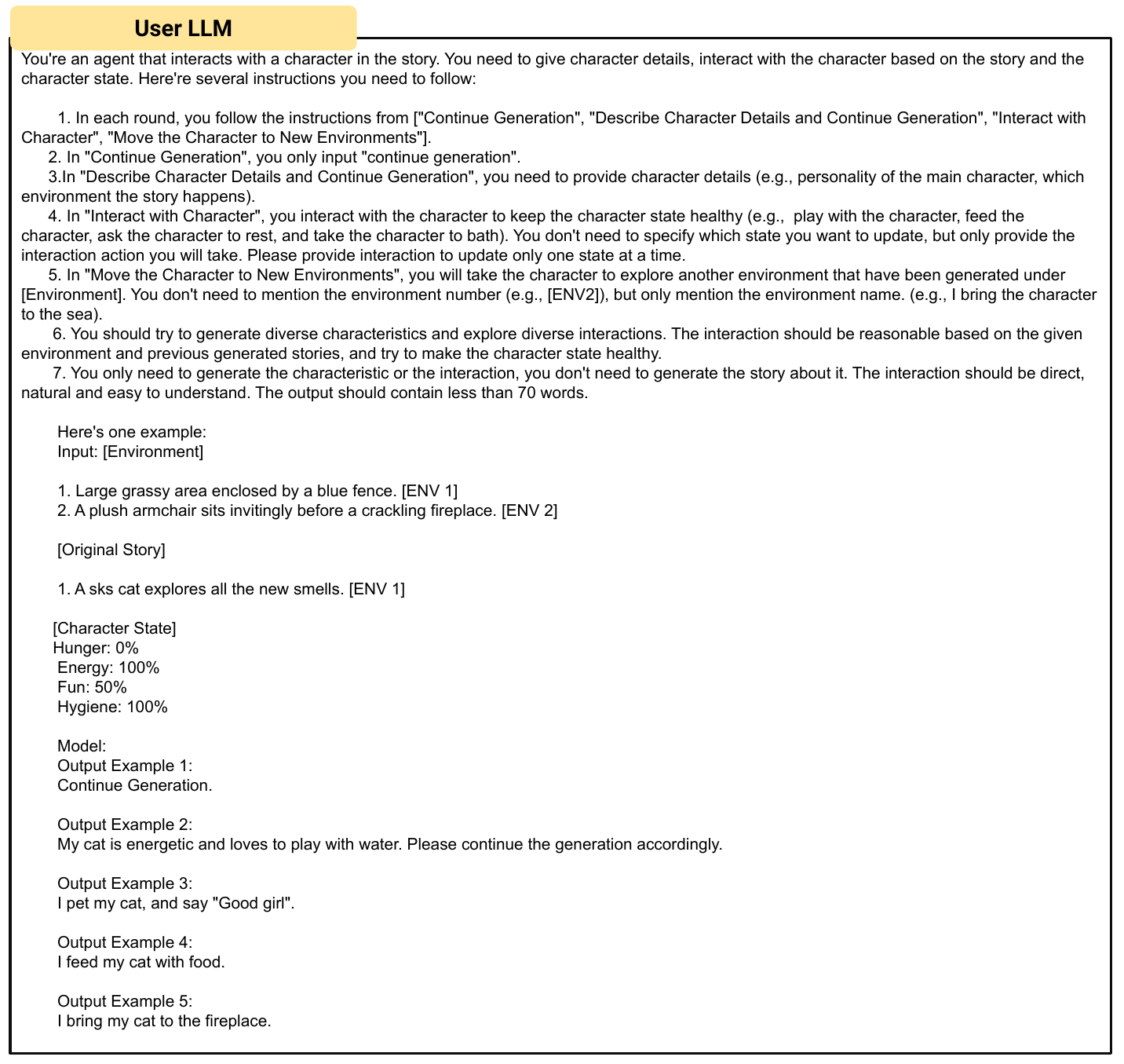}
    \caption{Prompt template used to query user LLM.
    }
   \label{fig:user_prompt}
\end{figure}

\begin{figure}[t!]
    \centering
    \includegraphics[width=\textwidth]{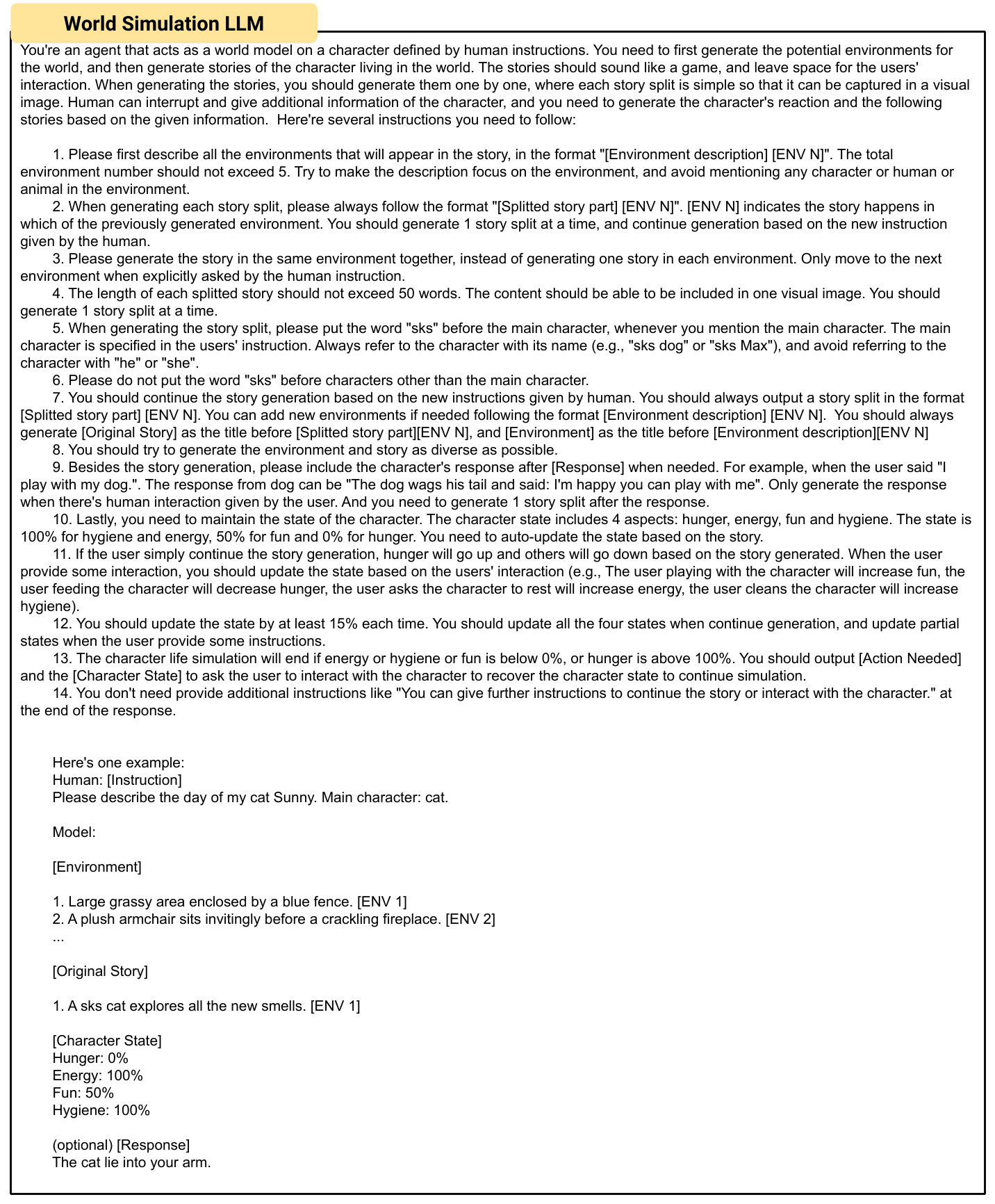}
    \caption{Prompt template used to query world simulation LLM.
    }
   \label{fig:world_prompt}
\end{figure}

In this section, we provide the prompt templates we use to collect the user-simulation data. Specifically, we query GPT-3.5 to generate diverse topics and character descriptions using the template shown in Figure~\ref{fig:topic_prompt}. The prompt template for guiding the potential user interactions for user LLM is shown in Figure~\ref{fig:user_prompt}. To ensure user interactions align with the dialogue history, we include the interaction history as input to the user LLM. The world simulation LLM prompt template, shown in Figure~\ref{fig:world_prompt}, also takes in the dialogue history and generates the next character actions, states and narratives. We constraint the world simulation LLM to generate one storyline at a time, allowing users to choose how to continue the story.

\section{Evaluation Prompt} \label{sec:llm_evaluation}

\begin{figure}[t!]
    \centering
    \includegraphics[width=\textwidth]{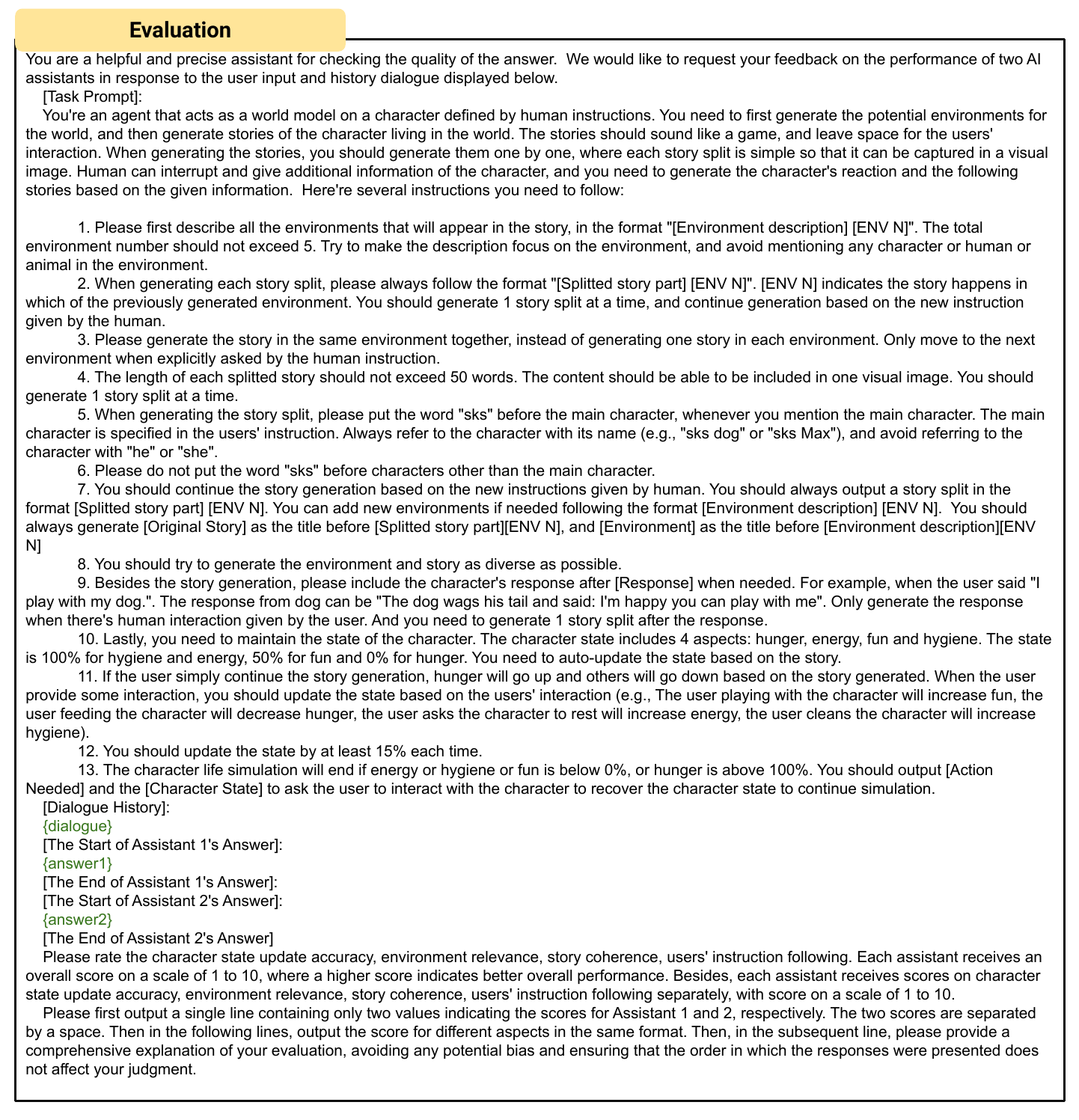}
    \caption{Prompt template used to query GPT4 to compare the performance between two LLM outputs. The prompt is adapted from Vicuna~\citep{vicuna2023}.
    }
   \label{fig:gpt_eval}
\end{figure}

We include the prompt template we use to compare the outputs from two LLMs in Figure~\ref{fig:gpt_eval}. The prompt is adapted from Vicuna~\citep{vicuna2023}, and has been validated as an effective tool for comparing the performance of two LLMs on a given task.

\section{Reproducibility Statement} \label{sec:reproduce}

First, we include the implementation details of \methodname{} in Sec.~\ref{sec:implementation}, covering the training hyperparameters for Dreambooth fine-tuning, and LLM distillation, and the hyperparameters we use during inference for both LLM and image generation. Second, we provide detailed description of the user-simulation data we collect for training in Sec.~\ref{sec:lang}, and further include the prompt template used to query GPT models in Appendix Sec.~\ref{sec:data_template}. Lastly, for the LLM-based evaluation, the prompt template for querying GPT-4 to compare two LLM outputs is provided in Appendix Sec.~\ref{sec:llm_evaluation}.

\end{document}